\documentclass{article}

\usepackage{microtype}
\usepackage{graphicx}
\usepackage{subfigure}
\usepackage{comment}
\usepackage{booktabs} 

\usepackage{hyperref}

\usepackage{amsmath}
\usepackage{amssymb}
\usepackage{mathtools}
\usepackage{amsthm}
\usepackage{bm} 
\usepackage[normalem]{ulem}

\usepackage[capitalize,noabbrev]{cleveref}

\def\R{\mathbb{R}}
\def\C{\mathbb{C}}
\def\Z{\mathbb{Z}}

\def\msets{{\rm msets}}

\def\rr{{\boldsymbol{r}}}
\def\pp{{\boldsymbol{p}}}
\def\kk{{\boldsymbol{k}}}
\def\AA{\boldsymbol{A}}
\def\BB{\boldsymbol{B}}
\def\xx{{\bm x}}

\def\balpha{{\boldsymbol{\alpha}}}

\def\calK{\mathcal{K}}
\def\calC{\mathcal{C}}
\theoremstyle{plain}
\newtheorem{theorem}{Theorem}[section]

\theoremstyle{definition}

\theoremstyle{remark}

\usepackage[utf8]{inputenc} 
\usepackage[T1]{fontenc}    
\usepackage{hyperref}       
\usepackage{url}            
\usepackage{booktabs}       
\usepackage{amsfonts}       
\usepackage{nicefrac}       
\usepackage{microtype}      
\usepackage{xcolor}         

\newcommand{\GL}{\mathrm{GL}}
\newcommand{\SL}{\mathrm{SL}}
\newcommand{\Sp}{\mathrm{Sp}}
\newcommand{\SO}{\mathrm{SO}}
\newcommand{\gO}{\mathrm{O}}
\newcommand{\SU}{\mathrm{SU}}

\newcommand{\EucGp}{\mathrm{E}(3)}

\DeclareMathOperator{\Sym}{Sym}
\DeclareMathOperator{\Lie}{Lie}
\newcommand\lieg{\mathfrak{g}}
\newcommand\lieh{\mathfrak{h}}

\definecolor{airforce}{rgb}{0.36, 0.54, 0.86}
\usepackage[textsize=tiny]{todonotes}

\usepackage[preprint]{neurips_2023}

\title{A General Framework for Equivariant Neural Networks on Reductive Lie Groups}

\author{%
  Ilyes Batatia \\
  Engineering Laboratory, \\
  University of Cambridge\\
  Cambridge, CB2 1PZ UK \\
  Department of Chemistry, \\
  ENS Paris-Saclay,
  Université Paris-Saclay \\
  91190 Gif-sur-Yvette, France \\
  \texttt{ilyes.batatia@ens-paris-saclay.fr} \\
  \And
   Mario Geiger \\
   Department of Electrical Engineering \\
   and Computer Science, \\
   Massachusetts Institute of Technology \\
   Cambridge, MA, USA \\
   \AND
   Jose Munoz \\
   EIA University, FTA Group \\
   Antioquia, Colombia \\
   \And
   Tess Smidt \\
   Department of Electrical Engineering \\
   and Computer Science, \\
   Massachusetts Institute of Technology \\
   Cambridge, MA, USA \\
   \And
   Lior Silberman \\  
   Department of Mathematics \\
  University of British Columbia \\
  Vancouver, BC, Canada V6T 1Z2 \\
  \And
  Christoph Ortner \\
  Department of Mathematics \\
  University of British Columbia \\
  Vancouver, BC, Canada V6T 1Z2 \\
  \texttt{} \\
}

\begin{document}

\maketitle

\vspace{-0.6cm}

\begin{abstract}
 Reductive Lie Groups, such as the orthogonal groups, the Lorentz group, or the unitary groups, play essential roles across scientific fields as diverse as high energy physics, quantum mechanics, quantum chromodynamics, molecular dynamics, computer vision, and imaging. In this paper, we present a general Equivariant Neural Network architecture capable of respecting the symmetries of the finite-dimensional representations of any reductive Lie Group $G$. Our approach generalizes the successful ACE and MACE architectures for atomistic point clouds to any data equivariant to a reductive Lie group action. We also introduce the {\tt lie-nn} software library, which provides all the necessary tools to develop and implement such general $G$-equivariant neural networks. It implements routines for the reduction of generic tensor products of representations into irreducible representations, making it easy to apply our architecture to a wide range of problems and groups. The generality and performance of our approach are demonstrated by applying it to the tasks of top quark decay tagging (Lorentz group) and shape recognition (orthogonal group).
 \end{abstract}
\section{Introduction}
\vspace{-3mm}
Convolutional Neural Networks (CNNs) \citep{CNNlecun} have become a widely used and powerful tool for computer vision tasks, in large part due to their ability to achieve translation equivariance. This property led to improved generalization and a significant reduction in the number of parameters.
Translation equivariance is one of many possible symmetries occurring in machine learning tasks.

A wide range of symmetries described by reductive Lie Groups is present in physics, such as $O(3)$ in molecular mechanics, $\SO(1,3)$ in High-Energy Physics, $\SU(2^{N})$  
in quantum mechanics, and $\SU(3)$ in quantum chromodynamics. 
Machine learning architectures that respect these symmetries often lead to significantly improved predictions while requiring far less training data. This has been demonstrated in many applications including 2D imaging with $\gO(2)$ symmetry~\citep{CohenSteerable2016, estevespolar2017}, machine learning force fields with $\gO(3)$ symmetry~\citep{Anderson2019CormorantCM, Bartok2013OnEnvironments, nequip, Batatia2022mace} or jet tagging with $\SO^+(1,3)$ symmetry~\citep{Bogatskiy:2022czk, Li:2022xfc}.

One way to extend CNNs to other groups \citep{pmlr-v119-finzi20a, pmlr-v80-kondor18a} is through harmonic analysis on homogeneous spaces, where the convolution becomes an integral over the group. Other architectures work directly with finite-dimensional representations. We follow the demonstration of \cite{Bogatskiy:2020tje} who constructed a universal approximation of any equivariant map with a feed-forward neural network with vector activations belonging to finite-dimensional representations of a wide class of Lie groups.
In this way, one can avoid computational challenges created by infinite-dimensional representations.
\begin{figure}[h]
    \centering
    \includegraphics[width=0.6\textwidth]{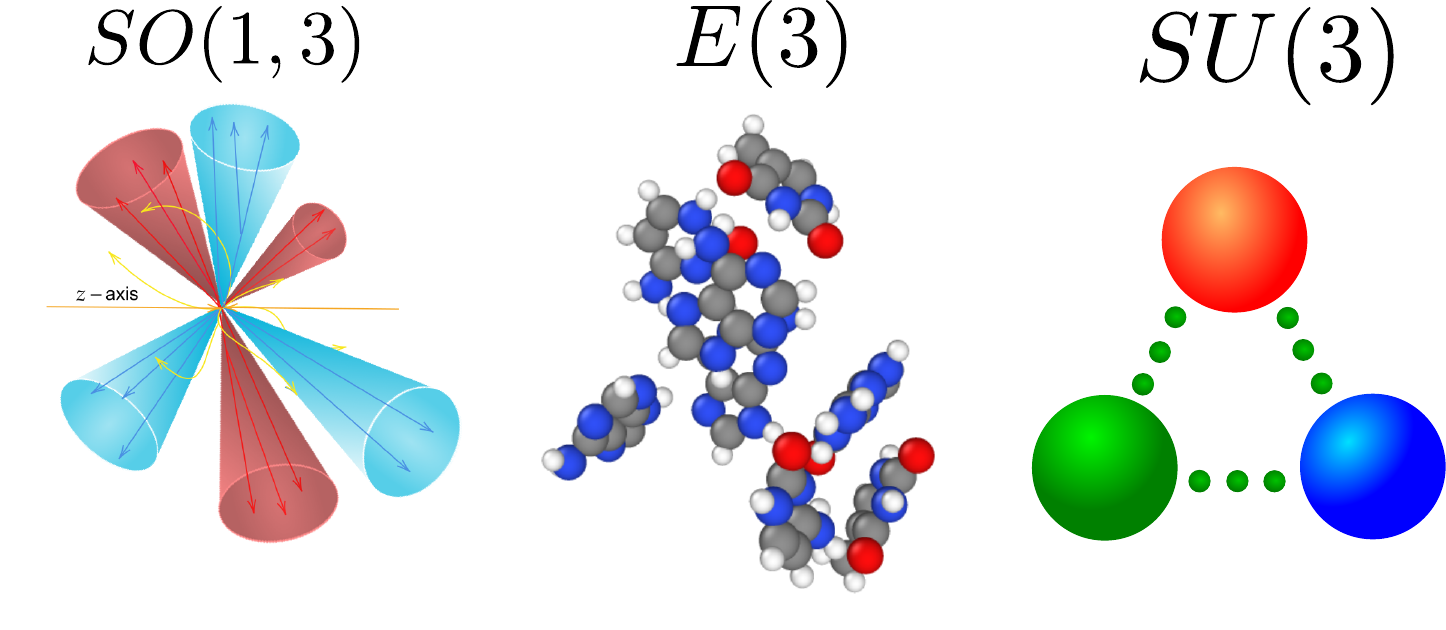}
    \caption{Examples of natural science problems and associated reductive Lie groups. For high energy physics, the Lorentz group $\SO(1,3)$; for chemistry, the Euclidean group $\EucGp$; for quantum-chromodynamics, the $\SU(3)$ group.}
    \label{fig:physics_overview}
\end{figure}

Alternatively, our current work can be thought of as a generalization of the Atomic Cluster Expansion (ACE) formalism of \cite{PhysRevB.99.014104} to general Lie groups. The ACE formalism provides a complete body-ordered basis of $\gO(3)$-invariant features. By combining the concepts of ACE and $\EucGp$-equivariant neural networks, \cite{Batatia2022mace} proposed the MACE architecture, which achieves state-of-the-art performance on learning tasks in molecular modelling. The present work generalizes the ACE and MACE architectures to arbitrary Lie groups in order to propose a generic architecture for creating representations of geometric point clouds in interaction. 

Concretely, our work makes the following contributions: \\[-6mm]
\begin{itemize} \setlength\itemsep{0mm}
    \item We develop the $G$-Equivariant Cluster Expansion. This new framework generalizes the ACE~\citep{PhysRevB.99.014104} and MACE~\citep{Batatia2022de} architectures to parameterize properties of point clouds that are equivariant under the action of a reductive Lie group $G$. 
    \item We prove that our architecture is universal, even for a single layer. 
    \item We introduce \texttt{lie-nn}, a new library providing all the essential tools to apply our framework to a variety of essential Lie Groups in physics and computer visions, including the Lorentz group, $\SU(N)$, $\SL_2(\C)$ and product groups.
    \item We demonstrate the generality and efficiency of our general-purpose approach by demonstrating excellent accuracy on two prototype applications, jet tagging, and 3D point cloud recognition.
\end{itemize}

\section{Background}
\vspace{-0.3cm}
We briefly review a few important group-theoretic concepts: 
A real (complex) {\bf Lie group} is a group that is also a finite-dimensional smooth (complex) manifold in which the product and inversion of the group are also smooth (holomorphic) maps. Among the most important Lie groups are Matrix Lie groups, which are closed subgroups of $\GL(n,\C)$ the group of invertible $n\times n$ matrices with complex entries. This includes well-known groups such as $\Sp(2n, \R)$ consisting of matrices of determinant one, that is relevant in Hamiltonian dynamics 
A finite-dimensional {\bf representation} of the Lie group $G$ is a finite-dimensional vector space $V$ endowed with a smooth homomorphism $\rho \colon G \to \GL(V)$. 
Features in the equivariant neural networks live in these vector spaces. 
An {\bf irreducible} representation $V$ is a representation that has no subspaces which are invariant under the action of the group (other than $\{0\}$ and $V$ itself). 
This means that $V$ can not be decomposed non-trivially as the direct sum of representations.
A {\bf reductive group} over a field $F$ is a (Zariski-) closed subgroup of the group of matrices $\GL(n, F)$ such that every finite-dimensional representation of $G$ on an $F$-vectorspace can be decomposed as a sum of irreducible representations. 

\vspace{-0.2cm}
\section{Related Work}
\vspace{-0.30cm}

\paragraph{Lie group convolutions}
Convolutional neural networks (CNNs), which are translation equivariant, have also been  generalized to other symmetries. For example, G-convolutions \citep{pmlr-v48-cohenc16} generalized CNNs to discrete groups. Steerable CNNs \citep{CohenSteerable2016} generalized CNNs to $O(2)$ equivariance and Spherical CNNs \citep{s.2018spherical} $O(3)$ equivariance. A general theory of convolution on any compact group and symmetric space was given by \cite{pmlr-v80-kondor18a}. This work was further extended to equivariant convolutions on Riemannian manifolds by \cite{WeilerManifold2021}.
\vspace{-0.35cm}

\paragraph{ACE} 
The Atomic Cluster Expansion (ACE)~\citep{PhysRevB.99.014104} introduced a systematic framework for constructing complete $O(3)$-invariant high body order  basis sets with constant cost per basis function, independent of body order~\citep{DUSSON2022110946}.
\vspace{-0.32cm}

\paragraph{e3nn + Equivariant MLPs} The  e3nn library  \citep{Geiger:2022ede} provides a complete solution to build $E(3)-$equivariant neural networks based on irreducible representations. The Equivariant MLPs \citep{Finzi2021:mlp} include more groups, such as $SO(1,3)$, $Z_{n}$, but are restricted to reducible representations making them much less computationally efficient than irreducible representations. 
\vspace{-0.35cm}

\paragraph{Equivariant MPNNs and MACE} Equivariant MPNNs \citep{NEURIPS2018_a3fc981a, Anderson2019CormorantCM, Bogatskiy:2020tje, Welling2021EGNN, brandstetter2022geometric, nequip} have emerged as a powerful architecture to learn on geometric point clouds. They construct permutation invariants and group equivariant representations of point clouds. Successful applications include simulations in chemistry, particle physics, and 3D vision. MACE \citep{Batatia2022mace} generalized the $O(3)$-Equivariant MPNNs to build messages of arbitrary body order, outperforming other approaches on molecular tasks. \citep{Batatia2022de} showed that the MACE design space is large enough to include most of the previously published equivariant architectures.

\section{The $G$-Equivariant Cluster Expansion}
\vspace{-0.3cm}
We are concerned with the representation of properties of point clouds. Point clouds are described as multi-sets (unordered tuples) $X = [x_i]_i$ where each particle $x_i$ belongs to a configuration domain $\Omega$. We denote the set of all such multi-sets by $\msets(\Omega)$. For example, in molecular modeling, $x_i$ might describe the position and species of an atom and therefore $x_i = (\rr_i, Z_i) \in \R^3 \times \Z$, while in high energy physics, one commonly uses the four-momentum $x_i = (E_i, \pp_i) \in \R^4$, but one could also include additional features such as charge, spin, and so forth.

A property of the point cloud is a map
\begin{equation}
    \Phi \colon \msets(\Omega) \to Z
\end{equation}
i.e., $X \mapsto \Phi(X) \in Z$, 
usually a scalar or tensor. The range space $Z$ is application dependent and left abstract throughout this paper. 
Expressing the input as a multi-set implicitly entails two important facts: (1) it can have varying lengths; (2) it is invariant under the permutations of the particles. The  developed in this article are also applicable to fixed-length multi-sets, in which case $\Phi$ is simply a permutation-invariant function defined on some $\Omega^N$. Mappings that are not permutation-invariant are special case with several simplifications.

In many applications, especially in the natural sciences, particle properties satisfy additional symmetries. When a group $G$ acts on $\Omega$ as well as on $Z$ we say that $\Phi$ is $G$-\textbf{equivariant} if 
\begin{equation}
    \Phi \circ g =  \rho_Z(g) \Phi,\qquad g \in G
\end{equation}
where $\rho_Z (g)$ is the action of the group element $g$ on the range space $Z$.  In order to effectively incorporate exact group symmetry into properties $\Phi$, 
we consider model architectures of the form 
\begin{equation} 
    \label{eq:Phi_embedding}
    \Phi \colon {\rm msets}(\Omega) \underset{\text{\tiny embedding}}{\longrightarrow}
            V  \underset{\text{\tiny parameterization}}{\longrightarrow}
            V  \underset{\text{\tiny readout}}{\longrightarrow}
            Z, 
\end{equation} 
where the space $V$ into which we ``embed'' the parameterization is a possibly infinite-dimensional vector space in which a convenient representation of the group is available. For simplicity we will sometimes assume that $Z = V$.

The Atomic Cluster Expansion (ACE) framework~\citep{PhysRevB.99.014104, DUSSON2022110946,ACE_equivariant_ralf}) produces a complete linear basis for the space of all ``smooth'' $G$-equivariant properties $\Phi$ for the specific case when $G = \gO(3)$ and $x_i$ are vectorial interatomic distances. Aspects of the ACE framework were incorporated into $\EucGp$-equivariant message passing architectures, with significant improvements in accuracy~\citep{Batatia2022mace}. 
In the following paragraphs we demonstrate that these ideas readily generalize to arbitrary reductive Lie groups.

\vspace{-2mm}
\subsection{Efficient many-body expansion} 
\vspace{-2mm}
The first step is to expand $\Phi$ in terms of body orders, and truncate the expansion at a finite order $N$:  
\begin{equation}  \label{eq:manybody}
\begin{split} 
    \Phi^{(N)}(X) &= \varphi_0 + 
    \sum_i \varphi_1(x_i) + \sum_{i_1, i_2} \varphi_2(x_{i_1}, x_{i_2}) 
    + \dots + 
    \sum_{i_1, \dots, i_N} 
    \varphi_N(x_{i_1}, \dots, x_{i_N}), 
\end{split}
\end{equation}
where $\varphi_n$ defines the $n$-body interaction. Formally, the expansion becomes systematic in the limit as $N \to \infty$.  The second step is the expansion of the $n$-particle functions $\varphi_n$ in terms of a symmetrized tensor product basis. To define this we first need to specify the embedding of particles $x$: A countable family $(\phi_k)_{k}$ is a 1-particle basis if they are linearly independent on $\Omega$ and any smooth 1-particle function $\varphi_1$ (not necessarily equivariant) can be expanded in terms of $(\phi_k)_k$, i.e, 
\begin{equation} \label{eq:dense_1p}
    \varphi_1(x) = \sum_{k} w_{k} \phi_k(x).
\end{equation}
For the sake of concreteness, we assume that $\phi_k : \Omega \to \mathbb{C}$, but the range can in principle be any field. Let a complex vector space $V$ be given, into which the particle embedding maps, i.e., 
\[
    (\phi_k(x))_{k} \in V \qquad \forall x \in \Omega.
\] 
As a consequence of \eqref{eq:dense_1p} any smooth scalar $n$-particle function $\varphi_n$ can be expanded 
in terms of the corresponding tensor product basis, 
\begin{equation}
    \varphi_n(x_1, \dots, x_n) = \sum_{k_1, \dots, k_n} 
            w_{k_1\dots k_n} \prod_{s=1}^n \phi_{k_s}(x_s).
\end{equation}
Inserting these expansions into \eqref{eq:manybody} and interchanging summation (see appendix for the details) we arrive at a model for scalar permutation-symmetric properties, 
\begin{equation} \label{eq:CE}
\begin{split}
    A_k = \sum_{x \in X} \phi_k(x), \qquad
    \AA_{\kk} = \prod_{s = 1}^{n} A_k, \qquad
    \Phi^{(N)}= \sum_{\kk \in \calK} w_{\kk} \AA_{\kk},
\end{split}
\end{equation}
where $\calK$ is the set of all $\kk$ tuples indexing the features $\AA_{\kk}$. Since $\AA_{\kk}$ is invariant under permuting $\kk$, only ordered $\kk$ tuples are retained. 
The features $A_k$ are an embedding of ${\rm msets}(\Omega)$ into the space $V$. The tensorial product features (basis functions) $\AA_{\kk}$ form a complete linear basis of multi-set functions on $\Omega$ and the weights $w_{\kk}$ can be understood as a symmetric tensor. 
We will extend this linear cluster expansion model $\Phi^{(N)}$ to a message-passing type neural network model in \S~\ref{sec:mace}.

We remark that, while the standard tensor product embeds $(\otimes_{s=1}^n \phi_{k_s})_{\kk} \colon \Omega^n \to V^n$, the $n$-correlations $\AA_{\kk}$ are {\em symmetric tensors} and embed $(\AA_{\kk})_{\kk} \colon \msets(\Omega) \to \Sym^n V$.  

\vspace{-2mm} 

\subsection{Symmetrisation}
\vspace{-3mm}
With \eqref{eq:CE} we obtained a systematic linear model for (smooth) multi-set functions. It remains to incorporate $G$-equivariance. 
We assume that $G$ is a reductive Lie group with a locally finite representation in $V$. 
In other words we choose a representation $\rho = (\rho_{kk'})\colon G \to \GL(V)$ such that
\begin{equation} \label{eq:rep_indexnotation}
    \phi_k \circ g = \sum_{k'} \rho_{kk'}(g) \phi_{k'},
\end{equation}
where for each $k$ the sum over $k'$ is over a finite index-set depending only on $k$. 
Most Lie groups one encounters in physical applications belong to this class, the affine groups being notable exceptions. 
However, those can usually be treated in an {\it ad hoc} fashion, which is done in all $E(3)$-equivariant architectures we are aware of. 
In practice, these requirements restrict how we can choose the embedding $(\phi_k)_k$. If the point clouds $X = [x_i]_i$ are already given in terms of a representation of the group, then one may simply construct $V$ to be iterative tensor products of $\Omega$; see e.g. the MTP~\citep{Shapeev2016-pd} and PELICAN~\citep{Bogatskiy:2022czk} models.
To construct an equivariant two-particle basis we need to first construct the set of all intertwining operators from $V \otimes V \to V$. Concretely, we seek all solutions  $C_{k_1 k_2}^{\balpha,K}$ to the equation
\begin{equation}
    \label{eq:Clebsch--Gordan-2}
    \sum_{k_1' k_2'} C_{k_1' k_2'}^{\balpha,K} 
    \rho_{k_{1}'k_{1}}(g) \rho_{k_{2}' k_{2}}(g) = 
    \sum_{K'} \rho_{K K'}(g) C_{k_1 k_2}^{\balpha,K'};
\end{equation}
or, written in operator notation, 
\begin{equation}
    \label{eq:Clebsch--Gordan-2-op}
    C^\balpha \rho \otimes \rho = 
    \rho  C^\balpha.
\end{equation}
We will call the $C^{\balpha,K}_{\kk}$ \emph{generalized Clebsch--Gordan coefficients} since in the case $G=\SO(3)$ acting on the spherical harmonics embedding $\phi_{lm} = Y_l^m$ those coefficients are exactly the classical Clebsch--Gordan coefficients.
The index $\balpha$ enumerates a basis of the space of all solutions to this equation. For the most common groups, one normally identifies a canonical basis $C^\balpha$ and assigns a natural meaning to this index (cf. \S~\ref{appendix:labeling}).
Our abstract notation is chosen because of its generality and convenience for designing computational schemes. 
The generalization of the Clebsch--Gordan equation \eqref{eq:Clebsch--Gordan-2} to $n$ products of representations acting on the symmetric tensor space ${\rm Sym}^n(V)$ becomes (cf. \S~\ref{sec:symmetric_tp})
\begin{equation} \label{eq:generalized_cg_eqn_sym}
    \begin{split}
    &\sum_{\kk'} {\calC}^{\balpha, K}_{\kk'} \overline{\bm \rho}_{\kk' \kk} = 
    \sum_{K'} \rho_{K K'} {\calC}^{\balpha, K'}_{\kk}  \qquad 
    \forall K, \quad \kk = (k_1, \dots, k_N), \quad g \in G,  \\ 
    &\text{where} \qquad  \overline{\bm \rho}_{\kk' \kk} = 
    \sum_{\substack{\kk'' = \pi \kk' \\ \pi \in S_n}}{\bm \rho}_{\kk'' \kk}
    \qquad \text{and} \qquad 
    {\bm \rho}_{\kk' \kk} = \prod_{t = 1}^n \rho_{k_t' k_t}.
    \end{split}
\end{equation}
Due to the symmetry of the $(\AA_{\kk})_{\kk}$ tensors $\calC^{\balpha, K}_{\kk}$ need only be computed for ordered $\kk$ tuples and the sum $\sum_{\kk'}$ also runs only over ordered $\kk$ tuples.  Again, the index $\balpha$ enumerates a basis of the space of solutions. 
Equivalently, \eqref{eq:generalized_cg_eqn_sym} can be written in compact notation as 
\begin{equation}
    \calC^{\balpha} \overline{\bm \rho} = \rho \calC^{\balpha}. 
\end{equation}
These coupling operators for $N$ products can often (but not always) be constructed recursively from couplings of pairs \eqref{eq:Clebsch--Gordan-2}. 

We can now define the symmetrized basis 
\begin{equation}
\label{eq:b-basis-equivariant}
    \BB_{\balpha}^K = \sum_{\kk'} \calC^{\balpha,K}_{\kk'} \AA_{\kk'}.
\end{equation}
The equivariance of \eqref{eq:b-basis-equivariant} is easily verified by applying a transformation $g\in G$ to the input (cf \S~\ref{sec:equivariant_proof}).

{\bf Universality: }
In the limit as the correlation order $N \to \infty$, the features $(\BB_{\balpha}^K)_{K, \balpha}$ form a complete basis of smooth equivariant multi-set functions, in a sense that we make precise in Appendix~\ref{sec:completeness}. Any equivariant property $\Phi_V : \Omega \to V$ can be approximated by a linear model
\begin{equation}
    \Phi_V^K = \sum_{\balpha} c_{\balpha}^K B_{\balpha}^K,
\end{equation}
to within arbitrary accuracy by taking the number of terms in the linear combination to infinity.


\vspace{-2mm}
\subsection{Dimension Reduction}
\vspace{-3mm}
The tensor product of the cluster expansion in \eqref{eq:CE} is taken on all the indices of the one-particle basis. Unless the embedding $(\phi_k)_k$ is very low-dimensional it is often preferable to ``sketch'' this tensor product. For example, consider the canonical embedding of an atom $x_i = (\rr_i, Z_i)$, 
\[
    \phi_k(x_i) = \phi_{znlm}(x_i) = \delta_{zZ_i} R_{nl}(r_i) Y_l^m(\hat\rr_i).
\]
Only the $(lm)$ channels are involved in the representation of $\gO(3)$ hence there is considerable freedom in ``compressing'' the $(zn)$ channels. 

Following \cite{derbytrace} we construct a sketched $G$-equivariant cluster expansion: We endow the one-particle basis with an additional index $c$, referred to as the sketched channel, replacing the index $k$ with the index pair $(c, k)$, and renaming the embedding  $(\phi_{ck})_{c,k}$. In the case of three-dimensional particles one may,  for example, choose $c = (z, n)$. In general it is crucial that the representation remains in terms of the $\rho_{k,k'}$, that is, \eqref{eq:rep_indexnotation} becomes 
\begin{equation} \label{eq:rep_sketch} 
    \phi_{ck} \circ g = \sum_{k'} \rho_{kk'}(g) \phi_{ck'}.
\end{equation} 
Therefore, manipulating only the $c$ channel does not change any symmetry properties of the architecture. We can use this fact to admit a learnable embedding, 
\[
    \tilde{A}_{ck} = \sum_{c'} w_{cc'} \phi_{c' k}, 
\]
This mechanism is employed in numerous architectures to reduce the dimensionality of the embedding, but the approach taken in by \cite{derbytrace} and \cite{Batatia2022de} and followed here is the exact opposite: we allow many more learnable $c$ channels but then decouple them resulting in a much lower-dimensional basis of $n$-correlations, defined by 
\begin{equation} 
\label{eq:sketched}
\begin{split}
    %
    \tilde{\AA}_{c\kk} &= \prod_{t = 1}^{n} \Bigg( \sum_{c'}w_{cc'}\sum_{x \in X} \phi_{c'k_t}(x) \Bigg).
\end{split}
\end{equation}
%
The resulting symmetrized basis is then obtained by 
\begin{equation}
    \label{eq:trace_basis}
    \BB_{c\balpha}^K = \sum_{\kk'} C_{\kk'}^{\balpha, K} \tilde{\AA}_{c\kk'}.
\end{equation}
Following the terminology of \cite{derbytrace} we call this architecture the tensor-reduced ACE, or, $G$-TRACE. There are numerous natural variations on its construction, but for the sake of simplicity, we restrict our presentation to this one case.

{\bf Universality: }
Following the proof of \cite{derbytrace} one can readily see that the $G$-TRACE architecture inherits the universality of the cluster expansion, in  the limit of decoupled channels $\#c \to \infty$.  
A smooth equivariant property $\Phi$ may be approximated to within arbitrary accuracy by an expansion $\Phi^K(X) \approx \sum_{c, \balpha} c^K_{\balpha} \BB^{K}_{c, \balpha}(X)$. Since the embedding $\tilde{A}_{ck}$ is learnable, this is a {\em nonlinear model}. We refer to \S~\ref{sec:completeness} for the details.

\subsection{G-MACE, Multi-layer cluster expansion}
\label{sec:mace}
The $G$-equivariant cluster expansion is readily generalized to a multi-layer architecture by re-expanding previous features in a new cluster expansion \citep{Batatia2022de}.
The multi-set $X$ is endowed with extra features, ${\bm h}_{i}^{t} = (h_{i,cK}^t)_{c,K}$, that are updated for $t \in \{1,...,T\}$ iterations. These features themselves are chosen to be a field of representations such that they have a well-defined transformation under the action of the group. This results in 
\begin{align}
\label{eq:state_multi}
    x_i^{t} &= (x_i, {\bm h}_{i}^{t}) \\ 
    \phi^t_{ck}(x_i, {\bm h}_i^t) &= \sum_{\balpha} w^{t, c k}_{\balpha} 
    \sum_{k', k''}
      C^{\balpha,k}_{k' k''} h_{i,c k'}^t \phi_{c k''}(x_i) 
\end{align}
The recursive update of the features proceeds as in a standard message-passing framework but with the unique aspect that messages are formed via the $G$-TRACE and in particular can contain arbitrary high correlation order:.
\begin{equation}
   m_{i,cK}^{t} = \sum_{\balpha} W_{\balpha}^{t,cK} \BB_{c\balpha}^{t,K}.
\end{equation}
The gathered message ${\bm m}_{i}^{t} = (m_{i,cK}^t)_{c,k}$ is then used to update the particle states,
\begin{equation}
\label{eq:state_update}
 x_i^{t+1} = (x_i, {\bm h}_{i}^{t+1}), \qquad {\bm h}_i^{t+1} = U_{t}\big({\bm m}_{i}^{t}\big), 
\end{equation}
where $U_t$ can be an arbitary fixed or learnable transformation (even the identity).
Lastly, a readout function maps the state of a particle to a target quantity of interest, which could be {\em local} to each particle or {\em global} to the mset $X$, 
\begin{equation}
\label{eq:readout}
y_{i} = \sum_{t=1}^{T} \mathcal{R}_t^{\rm loc}(x_i^{t}), \qquad \text{respectively,} 
\qquad y = \sum_{t=1}^{T} \mathcal{R}_t^{\rm glob}(\{x_i^{t}\}_{i}).
\end{equation}
This multi-layer architecture corresponds to a general message-passing neural network with arbitrary body order of the message at each layer. We will refer to this architecture as $G$-MACE. The $G$-MACE architecture directly inherits universality from the $G$-ACE and $G$-TRACE architectures:

\begin{theorem}[Universality of $G$-MACE]
    \label{th:universality}
    Assume that the one-particle embedding $(\phi_k)_k$ is a complete basis. Then, the set of $G$-MACE models, with a fixed finite number of layers $T$, is dense in the set of continuous and equivariant properties of point clouds $X \in \msets(\Omega)$, in the topology of pointwise convergence. It is dense in the uniform topology on compact and size-bounded subsets. 
\end{theorem}

\section{{\tt lie-nn} : Generating Irreducible Representations for Reductive Lie Groups}
\label{sec:genirreps}
\vspace{-0.3cm}
In order to construct the G-cluster expansion for arbitrary Lie groups, one needs to compute the generalized Clebsch--Gordan coefficients \eqref{eq:generalized_cg_eqn_sym} for a given tuple of representations (see~\ref{eq:b-basis-equivariant}). To facilitate this task, we have implemented an open source software library, \texttt{lie-nn}\footnote{{\tt https://github.com/lie-nn/lie-nn}}. In this section we review the key techniques employed in this library.

\subsection{Lie Algebras of Reductive Lie Groups}
\vspace{-0.2cm}
Formally, the Lie algebra of a Lie group is its tangent space at the origin and carries an additional structure, the Lie bracket.  Informally the Lie algebra can be thought of as a linear approximation to the Lie group but, due to the group structure, this linear approximation carries (almost) full information about the group.  In particular the representation theory of the Group is almost entirely determined by the Lie algebra, which is a simpler object to work with instead of the fully nonlinear Lie group.
\vspace{-0.3cm}
\paragraph{Lie algebra}
The Lie groups we study can be realized as closed subgroups $G\subset \GL_n(\R)$ of the general linear group.  In that case their Lie algebras can be concretely realized as $\lieg = \Lie(G) = \{ X\in M_n(\R) \mid \forall t\in\R: \exp(tX)\in G\}$ where $\exp(X) = 1 + X + \frac{1}{2} X^{2} ...$ is the standard matrix exponential.  It turns out that $\lieg \subset M_n(\R)$ is a linear subspace closed under the commutator bracket $[X,Y]=XY-YX$.
\paragraph{Structure theory}
We fix a linear basis $\{X_{i}\} \subset \lieg$, called a set of generators for the group.  The Lie algebra structure is determined by the \emph{structure constants} $A_{ijk}$ defined by $[X_{i}, X_{j}] = \sum_k A_{ijk} X_{k}$, in that if $X = \sum_i a_i X_i$ and $Y = \sum_j b_j X_j$ then $[X,Y] = \sum_k \left(\sum_{i,j} A_{ijk} a_i b_j\right) X_k$.
The classification of reductive groups provides convenient generating sets for their Lie algebras (or their complexifications).  One identifies a large commutative subalgebra $\lieh\subset\lieg$ (sometimes of $\lieg_\C = \lieg \otimes_\R \C$) with basis $\{H_i \}$ so that most (or all) of the other generators $E_\alpha$ can be chosen so that $[H_i,E_\alpha] = \alpha(H_i)E_\alpha$ for a linear function $\alpha$ on $\lieh$.  These functions are the so-called \emph{roots} of $\lieg$.  Structural information about $\lieg$ is commonly encoded pictorially via the \emph{Dynkin diagram} of $\lieg$, a finite graph the nodes of which are a certain subset of the roots.  There are four infinite families of simple complex Lie algebras $A_{n} = \mathfrak{su}(n+1), B_{n} =  \mathfrak{so}(2n+1), C_{n} =  \mathfrak{sp}(2n), D_{n} = \mathfrak{so}(2n)$ and further five exceptional simple complex Lie algebras (a general reductive Lie algebra is the direct sum of several simple ones and its centre). 
The Lie algebra only depends on the connected component of $G$.  thus when the group $G$ is disconnected in addition to the infinitesimal generators $\{X_i\}$ one also needs to fix so-called "discrete generators", a subset $\mathbf{H}\subset G$ containing a representative from each connected component.
\begin{figure}[ht]
    \centering
    \includegraphics[width=0.30\textwidth]{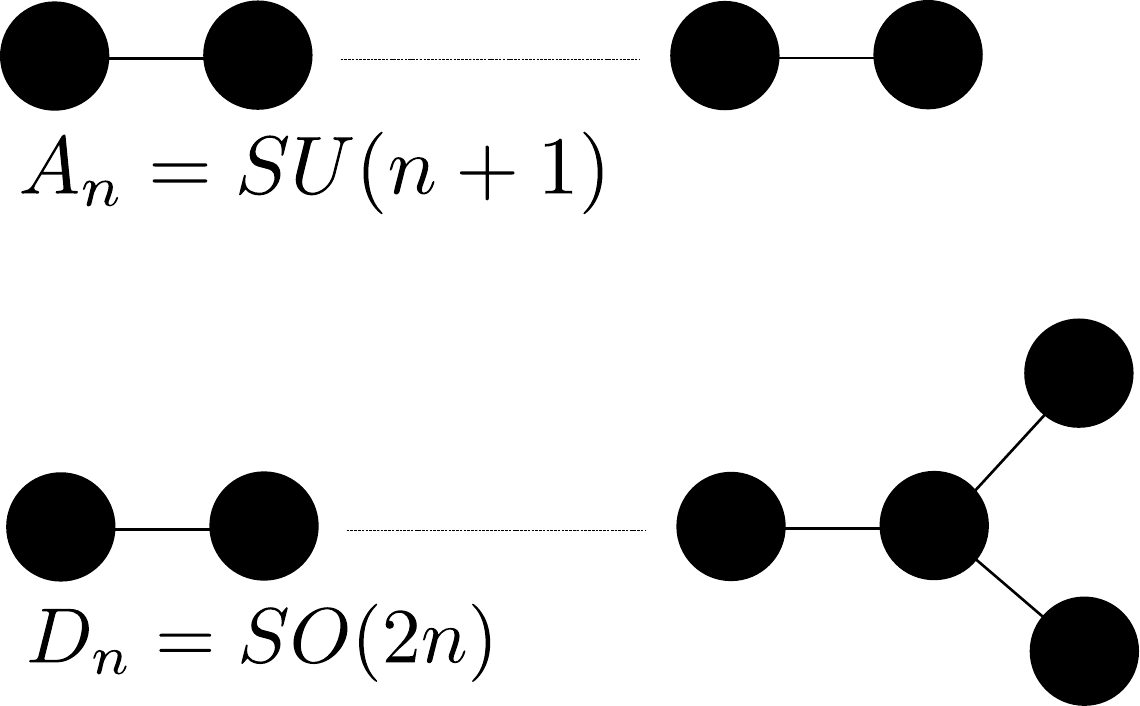}
    \caption{Examples of Dynkin diagrams and their associated group class.}
    \label{fig:dynkin_diagram}
\end{figure}
\vspace{-0.3pt}
\paragraph{Representation theory}
The representation theory of complex reductive Lie algebras is completely understood.  Every finite-dimensional representation is (isomorphic to) the direct sum of irreducible representations ("irreps"), with the latter parametrized by appropriate linear functional on $\lieh$ ("highest weight").  Further given a highest weight $\lambda$ there is a construction of the associated irrep with an explicit action of the infinitesimal generators chosen above.  The \textbf{Weyl Dimension Formula} gives the dimension of an irrep in terms of its highest weight.
\vspace{-10pt}
\subsection{Numerical Computations in \texttt{lie-nn}}
\vspace{-2mm}
The most basic class of the \texttt{lie-nn} library encodes a group $G$ and infinitesimal representation $d\rho$ of $\lieg$ using the tuple
\begin{equation}\label{eq:lie-nn-implement-rep}
    \rho := \left(A,n,\{d\rho(X_i)\}_{i},\{\rho(h)\}_{h\in\mathbf{H}}\right)\,,
\end{equation}

with $A$ the structure constants of the group, $n$ the dimension of the representation, and $d\rho(X_i)$ and $\rho(h)$ being $n\times n$ matrices encoding the action of the infinitesimal and the discrete generators respectively. The action of infinitesimal generators is related to the action of group generators by the exponential, $\forall X \in \lieg, \rho(e^{X}) = e^{d\rho(X)}$.

As the building blocks of the theory irreps are treated specially; the package implements functionality for the following operations for each supported Lie group:
\vspace{-5pt}
\begin{itemize} \setlength\itemsep{0mm}
    \item Constructing the irrep with a given highest weight.
    \item Determining the dimension of an irrep.
    \item Decomposing the tensor product of several irreps into irreps up to isomorphism (the \textbf{selection rule}, giving the list of irreducible components and their multiplicities).
    \item Decomposing the tensor product of several irreps into irreps explicitly via a change of basis ("generalized \textbf{Clebsch--Gordan} coefficients").
    \item Computating the symmetrized tensor product of the group (see. \ref{sec:symmetric_tp} for details).
\end{itemize}

To construct an irrep explicitly as in \eqref{eq:lie-nn-implement-rep} one needs to choose a basis in the abstract representation space (including a labeling scheme for the basis) so that we can give matrix representations for the action of generators. For this purpose, we use in \texttt{lie-nn} the Gelfand-Tsetlin (GT) basis \citep{gelfand1950:rsd} and associated labeling of the basis by GT patterns (this formalism was initially introduced for algebras of type $A_{n}$ but later generalized to all classical groups). Enumerating the GT patterns for a given algebra gives the dimension of a given irrep, the selection rules can be determined combinatorially, and it is also possible to give explicit algorithms to compute Clebsch--Gordan coefficients (the case of $A_{n}$ is treated by \cite{Alex_2011}). 
For some specific groups, simplifications to this procedure are possible and GT patterns are not required.

In some cases, one wants to compute coefficients for reducible representations or for representations where the analytical computation with GT patterns is too complex. In these cases, a numerical algorithm to compute the coefficients is required.  
Let $d\rho_{1}, d\rho_{2}$ be two Lie aglebra representations of interest.
The tensor product on the Lie algebra $d\rho_{1}\otimes d\rho_{2}(X)$ can be computed as,
\begin{align}
    d\rho_{1} \otimes d\rho_{2}\;(X) = d\rho_{1}(X) \otimes 1 + 1 \otimes d\rho_{2}(X)
\end{align}
Therefore, given sets of generators of three representations $d\rho_{1}, d\rho_{2}, d\rho_{3}$, the Clebsch--Gordan coefficients are the change of basis between $(d\rho_{1}(X) \otimes 1 + 1 \otimes d\rho_{2}(X))$ and $d\rho_{3}(X)$.
One can compute this change of basis numerically via a null space algorithm. For some groups, one can apply an iterative algorithm that generates all irreps starting with a single representation, using the above-mentioned procedure (see \ref{sec:iterative}).

\vspace{-0.2cm}

\section{Applications}
\vspace{-2mm}

\subsection{Lie groups and their applications}
\vspace{-2mm}
In Table \ref{tab:applications} we give a non-exhaustive overview of Lie groups and their typical application domains, to which our methodology naturally applies.

\begin{table}[h!]
    \centering
    \label{tab:applications}
    \caption{Lie groups of interests covered by the present methods and their potential applications to equivariant neural networks. The groups above the horizontal line are already available in \texttt{lie-nn}. The ones below the line fall within our framework and can be added.}
    \scalebox{0.99}{
    \begin{tabular}{l r r}
    \toprule 
     Group & Application & Reference  \\
    \midrule
     $\SU(1)$ & Electromagnetism &  \small{\citep{10.1007/978-3-030-80209-7_62}} \\
     $\SU(3)$ & Quantum Chromodynamics &  \small{\citep{latticefavoni2022}}\\
     $\SO(3)$ & 3D point clouds & \citep{Batatia2022mace}  \\
     $\SO^+(1,3)$ & Particle Physics & \small{\citep{pmlr-v119-bogatskiy20a}} \\
     $\SL(3, \R)$ & Point cloud classification & - \\
     $\SU(2^{N})$ & Entangled QP & - \\

     \midrule 
     $\Sp(N)$ & Hamiltonian dynamics &  - \\
     $\SO(2N + 1)$ & Projective geometry &  - \\
    \bottomrule
    \end{tabular}
    }
\end{table}

Benchmarking our method on all of these applications is beyond the scope of the present work, in particular, because most of these fields do not have standardized benchmarks and baselines to compare against. The MACE architecture has proven to be state of the art for a large range of atomistic modeling benchmarks \citep{Batatia2022mace}. In the next section, we choose two new prototypical applications and their respective groups to further assess the performance of our general approach.

\vspace{-0.2cm}
\subsection{Particle physics with the $SO(1,3)$}
\vspace{-0.2cm}
Jet tagging consists in identifying the process that generated a collimated spray of particles called a {\em jet} after a high-energy collision occurs at particle colliders. Each jet can be defined as a multiset of four-momenta $[ (E_i, \mathbf{p}_i) ]_{i=1}^N$, where $E_i\in \mathbb{R}^+ $ and $\mathbf{p}_i \in \mathbb{R}^3$.

Current state-of-the-art models incorporate the natural symmetry arising from relativistic objects, e.g, the Lorentz symmetry, as model invariance.
To showcase the performance and generality of the $G$-MACE framework we use the Top-Tagging dataset \citep{Kasieczka:2019dbj}, where the task is to differentiate boosted top quarks from the background composed of gluons and light quark jets. 
$G$-MACE achieves excellent accuracy, being the only arbitrary equivariant model to reach similar accuracy as PELICAN. We refer to Appendix~\ref{sec:jet_tagging_appendix}
for the details of the architecture. 

\begin{table}[h!]
\begin{center}
    \scalebox{.99}{
\begin{tabular}{lcccr}
\toprule
\midrule
Architecture & \#Params & Accuracy & AUC & Rej$_{30\%}$\\
\midrule \midrule
\textbf{PELICAN} & 45k & \textbf{0.942} & \textbf{0.987} & {$\bm{2289 \pm 204}$} \\
\textbf{partT}   & 2.14M & 0.940 & 0.986 & $1602 \pm 81$ \\
\textbf{ParticleNet} & 498k & 0.938 & 0.985 & $1298\pm46$\\
\textbf{LorentzNet} & 224k & \textbf{0.942} & \textbf{0.987} &  $2195\pm 173$\\
\textbf{BIP} &  {4k} & 0.931 & 0.981 & $853 \pm 68$ \\
\textbf{LGN}   &  {4.5k} & 0.929 & 0.964 & $435 \pm 95$ \\
\textbf{EFN}    &  82k & 0.927 & 0.979 & $888\pm 17$ \\
\textbf{TopoDNN}  &  59k & 0.916 & 0.972 & $295 \pm 5$\\

\textbf{LorentzMACE} & 228k & \textbf{0.942} & \textbf{0.987} & $1935 \pm 85 $ \\
\bottomrule
\end{tabular}
}
\end{center}
\caption{Comparisson between state-of-the-art metrics on the Top-Tagging dataset. Scores were taken from \citep{Bogatskiy:2022czk, Qu:2022mxj, Qu:2019gqs, Munoz:2022gjq, Bogatskiy:2020tje, Komiske:2018cqr, Pearkes:2017hku}.}
\end{table}

\vspace{-0.2cm}

\subsection{3D Shape recognition}
\vspace{-0.2cm}
3D shape recognition from point clouds is of central importance for computer vision. We use the ModelNet10 dataset \citep{WU2015:modelnet} to test our proposed architecture in this setting. As rotated objects need to map to the same class, we use a MACE model with $O(3)$ symmetry. 
To create an encoder version of $G$-MACE, we augment a PointNet++ implementation \citep{Pytorch_Pointnet} with $G$-MACE layers. See the appendix~\ref{sec:point_appendix} for more details on the architecture.

\begin{table}[h!]
    \centering
    \caption{Accuracy in shape recognition.}
    \scalebox{.99}{
    \begin{tabular}{l r r}
    \toprule
     Architecture &  & Accuracy  \\
    \midrule
     PointNet \citep{qi2016pointnet}  & & 94.2 \\
     PointNet ++ \citep{qi2017pointnetplusplus} & & 95.0 \\
     PointMACE (ours) & & \textbf{96.1} \\
    \bottomrule
    \end{tabular}
    }
\end{table}

\section{Conclusion}
\vspace{-0.2cm}
We introduced the $G$-Equivariant Cluster Expansion, which generalizes the successful ACE and MACE architectures to symmetries under arbitrary reductive Lie groups. We provide an open-source Python library \texttt{lie-nn} that provides all the essential tools to construct such general Lie-group equivariant neural networks. We demonstrated that the general $G$-MACE architecture simultaneously achieves excellent accuracy in Chemistry, Particle Physics, and Computer Vision. Future development will implement additional groups and generalize to new application domains. 

\begin{ack}
IB's work was supported by the ENS Paris Saclay. CO's work was supported by NSERC Discovery Grant IDGR019381 and NFRF Exploration Grant GR022937.
This work
was also performed using resources provided by the
Cambridge Service for Data Driven Discovery (CSD3).IB would like to thank Gábor Csányi for his support.

\end{ack}

\bibliography{main}

\newpage
\appendix

\section{Appendix}

\subsection{Proof of \eqref{eq:CE}}
This statement follows closely the arguments by \cite{DUSSON2022110946, ACE_equivariant_ralf} and others. 

\begin{align*}
    \sum_{j_1, \dots, j_n} \sum_{\kk} w_{\kk} \prod_s \phi_{k_s}(x_{j_s})
    &=
    \sum_{\kk} w_{\kk}  \sum_{j_1, \dots, j_n} \prod_s \phi_{k_s}(x_{j_s}) \\ 
    &= 
    \sum_{\kk} w_{\kk} \prod_{s = 1}^n \sum_j \phi_{k_s}(x_j) \\ 
    &= 
    \sum_{\kk} w_{\kk} \prod_{s = 1}^n A_k \\ 
    &= 
    \sum_{\kk} w_{\kk} {\bm A}_{\kk}. \\ 
\end{align*}

\subsection{Custom notation and indexing}
\label{appendix:labeling}
We briefly contrast our notation for Clebsch--Gordan coefficients \eqref{eq:generalized_cg_eqn_sym} with the standard notation. By means of example, consider the group $SO(3)$ in which case the Clebsch--Gordan equations are written as 

\begin{equation}
    \label{eq:Clebsch--Gordan-2-SO3}
    \begin{split}
    \sum_{m_1' m_2'} C_{l_1 m_1' l_2 m_2'}^{LM}
    \rho_{m_1' m_1}^{l_1}(g) \rho^{l_2}_{m_2' m_2}(g) 
    = 
    \sum_{M'} \rho^L_{M M'}(g) C_{l_1 m_1 l_2 m_2}^{L M'}.
    \end{split}
\end{equation}

In this setting, our index $\balpha$ simply enumerates all possible such coefficients. 
One can often assign a natural meaning to this index, e.g., for the group $SO(3)$ it is given by the pair of angular quantum numbers $(l_1, l_2)$. Specifically, in this case, we obtain 
\begin{equation} \label{eq:our_vs_classical_cgs}
    C^{\balpha, LM}_{l_1 m_1 l_2 m_2} = 
    \begin{cases}
       C^{LM}_{l_1 m_1 l_2 m_2}, & \text{if $\balpha = (l_1, l_2)$}, \\ 
       0, &  \text{otherwise,}
    \end{cases}
\end{equation}
where $C^{LM}_{l_1 m_1 l_2 m_2}$ are the Clebsch--Gordan coefficients in the classical notation.
Thus, the additional index $\balpha$ is not really required in the case of $SO(3)$, nor our other main example, $SO(1,3)$. Our notation is still useful to organize the computations of equivariant models, especially when additional channels are present, which is usually the case. Moreover, it allows for easy generalization to other groups where such a simple identification is not possible~\citep{Steinberg1961}.

\subsection{Equivariance of G-cluster expansion}
\label{sec:equivariant_proof}

The equivariance of the G-cluster expansion is easily verified by applying a transformation $g$ to the input,
\begin{equation}
\label{eq:b-basis-equivariant--}
\begin{split}
\BB_{\balpha}^K \circ g &= \sum_{\kk} \calC^{\balpha,K}_{\kk} \AA_{\kk} \circ g \\
&= 
\sum_{\kk} \calC^{\balpha,K}_{\kk} \left( \sum_{\kk'} \prod_{t} \rho_{k_{t},k_{t}'}(g) \AA_{\kk'} \right) \\
&= 
\sum_{\kk'} \left( \sum_{\kk} \calC^{\balpha,K}_{\kk} \prod_{t} \rho_{k_{t},k_{t}'}(g) \right) \AA_{\kk'} \\
&= 
\sum_{\kk'} \left( \sum_{K'} \rho_{KK'}(g) \calC^{\balpha,K'}_{\kk'} \right) \AA_{\kk'}  \\
&= \sum_{K'} \rho_{KK'}(g) \BB^{K'}_{\balpha}.
\end{split}
\end{equation}

\subsection{Completeness of the basis and Universality of MACE }
\label{sec:completeness}

We explain in which sense the basis $\BB^K_{\balpha}$ is a complete basis, and briefly sketch how to prove this claim. The argument is contained almost entirely in \citep{DUSSON2022110946} and only requires a single modification, namely Step 3 below, using a classical argument from representation theory. We will therefore give only a very brief summary and explain that necessary change.

We start with an arbitrary equivariant property $\Phi^V$ embedded in $V$ where we have a representation, i.e. the actual target property is $\Phi$ is then given as a linear mapping from $V$ to $Z$. For technical reasons, we require that only finitely many entries $\Phi^V_K$ may be non-zero, but this is consistent with common usage. For example, if $G = O(3)$ and if $\Phi$ is a scalar, then $\Phi^V_0 = \Phi$, while all other $\Phi^V_{LM} \equiv 0$. If $\Phi$ is a covariant vector, then $\Phi^V_{LM}$ is non-zero if and only if $L = 1$; and so forth. For other groups, the labeling may differ but the principle remains the same. 

{\it 1. Convergence of the cluster expansion. } The first step in our parameterisation is to approximate $\Phi^V$ in terms of a truncated many-body expansion \eqref{eq:manybody}. It is highly application-dependent on how fast this expansion converges. 
Rigorous results in this direction in the context of learning interatomic potentials can be found in \citep{2021-apxsym,2021-tbmanybody}. 
A generic statement can be made if the number of input particles is limited by an upper bound, in which case the expansion becomes exact for a finite $N$. This case leads to the uniform density result stated in Theorem~\ref{th:universality}. We adopt this setting for the time  being and return to the pointwise convergence setting below. 

In the uniform convergence setting we also require that the domain $\Omega$ is compact. 

Throughout the remainder of this section we may therefore assume that an $N$ can be chosen as well as smooth components $\varphi^{(n)}$ such that the resulting model $\Phi^{V, N}$ approximates $\Phi^V$ to within a target accuracy $\epsilon$, 
\[
    |\Phi^{V,N}_K({\bm x}) - \Phi^V_K({\bm x})| 
    \leq \epsilon 
    \qquad \forall {\bm x} \in {\rm msets}(\Omega). 
\]

{\it 2. The density of the embedding. }
As already stated in the main text, if the components $\varphi^{(n)}_K$ are smooth, and the embedding $\{\phi_k\}_k$ is dense in the space of one-particle functions \eqref{eq:dense_1p} then it follows that the $\varphi^{(n)}_K$ can be expanded in terms of the tensor product basis $\phi_{\kk} := \otimes_{s = 1}^n \phi_{k_s}$ to within arbitrary accuracy. The precise statement is the following standard result of approximation theory: if ${\rm span} \{\phi_k\}_k$ are dense in $C(\Omega)$, then ${\rm span} \{ \phi_{\kk} \}_{\kk}$ are dense in $C(\Omega^n)$. That is, for any $\epsilon > 0$, there exist approximants $p^{(n)}_K$ such that 
\[
    \| \varphi^{(n)}_K - p^{(n)}_K \|_\infty \leq \epsilon. 
\]
{\it 3. The density of the symmetrized basis. } 
The next and crucial step is to show that, if the $\varphi^{(n)}_K$ are equivariant, then the $p^{(n)}_K$ may be chosen equivariant as well without loss of accuracy. If the group $G$ is compact then the representations $\rho$ can be chosen unitary~\citep{Broecker1985}. In that case, the argument from \citep{DUSSON2022110946} can be used almost verbatim: let 
\[
    \bar{p}^{(n)}(\xx) := \int_G \rho(g)^{-1} p^{(n)}(g \xx) \, H(dg), 
\]
where $H$ is the normalized Haar measure then $\bar{p}^{(n)}$ is equivariant by construction and 
\begin{align*}
    &\big| \varphi^{(n)}(\xx) - \bar{p}^{(n)}(\xx) \big|  \\ 
    &= 
    \bigg| \int_G \rho(g)^{-1} \Big(
            \varphi^{(n)}(g \xx) - p^{(n)}(g \xx) \Big) \, H(dg)
    \bigg|  \\ 
    &\leq 
    \int_G \Big|
            \varphi^{(n)}(g \xx) - p^{(n)}(g \xx) \Big| \, H(dg) \\ 
    &\leq \int_G \| \varphi^{(n)} - p^{(n)} \|_\infty \, H(dg) \leq \epsilon. 
\end{align*}

If the group is not compact, then one can apply ``Weyl's Unitary Trick'' 
(see \citep{Bourbaki1989}, Ch. 3): first, one complexifies the group (if it is real) and then constructs a maximal compact subgroup  $K_{\mathbb{C}}$ of the complexification. This new group $K$ will have the same representation as $G$ and in virtue of being compact, that representation may again be chosen unitary. Therefore, symmetrizing $p^{(n)}$ with respect to  $K_{\mathbb{C}}$ results in an approximant that is not only equivariant w.r.t.  $K_{\mathbb{C}}$ but also equivariant w.r.t. $G$. 

{\it 4. The density of the basis $\BB^K_\balpha$. } As the last step one can readily observe that the symmetrization and cluster expansion steps can be exchanged. I.e. first symmetrizing and then employing the steps \eqref{eq:CE} result in the same model. Letting $\epsilon \to 0$ in the foregoing argument while fixing the number of particles $\#\xx$ results in all errors vanishing. Note that this will in particular require taking $N \to \infty$.

{\it 5. Pointwise convergence. } To obtain density in the sense of pointwise convergence we first introduce the {\it canonical cluster expansion} without self-interacting terms
\[
    \Phi_K(\xx) = \sum_{n = 0}^\infty \sum_{j_1 < \dots < j_n} v_{K}^{(n)}(x_{j_1}, \dots x_{j_n}).
\]
The difference here is that the summation is only over genuine sub-clusters. Because of this restriction the series is finite for all multi-set inputs $\xx$. In other words, it converges in the pointwise sense. 

One can easily see that $v_n$ can be chosen (explicitly) to make this expansion exact. After truncating the expansion at finite $n \leq N$ and then expanding the potentials $v_K^{(n)}$ one can exactly transform the canonical cluster expansion into the self-interacting cluster expansion. This procedure is detailed in \citep{DUSSON2022110946,ACE_equivariant_ralf}. 

The arguments up to this point establish the claimed universality for the linear ACE model. The corresponding universality of the TRACE model follows immediately from \citep{derbytrace}. Since a single layer of the MACE model is a TRACE model, this completes the proof of Theorem~\ref{th:universality}. 

\subsection{Product of groups}
Let $G_{1}$ and $G_{2}$ be two reductive Lie groups, and form the direct product group $G_{1} \times G_{2}$. and $\rho_{1}$ be a associated irreducible representations then $\rho_{1} \otimes \rho_{2}$ is an irreducible representation of $G_{1} \times G_{2}$. One can then generate any Clebsch--Gordan coefficient of the product group $G_{1} \times G_{2}$ using the algorithm presented above. It is of particular interest in the case of the equivariant message passing networks on points clouds, where the group of interest is $G \times S_{n}$.

\subsection{Symmetric Tensor products}
\label{sec:symmetric_tp}
\label{sec:symm_ten_prod}
The permutation group is an important concept in the context of tensor products. It can be useful to focus on a subset of the full tensor product space that exhibits certain permutation equivariance. For example, the spherical harmonics are defined as the permutation-invariant part of a tensor product.

The symmetric tensor product can be thought of as a change of basis, or projector, from the tensor product to the symmetric part of the tensor product. In the case of a tensor product of correlation order four we have,
\begin{equation}
    S_\nu = B_{\nu; ijkl} x_i y_j z_k w_l
\end{equation}
where $B$ is the change of basis that satisfies:
\begin{equation}
    B_{\nu; ijkl} = B_{\nu; \sigma(ijkl)} \forall \sigma \in S_4
\end{equation}
We propose in \texttt{lie-nn} a new algorithm used to calculate $B$. The Symmetric Tensor Product is calculated using a tree structure, starting at the leaves and progressing towards the trunk. The leaves are the basis of the individual indices, and they are combined and constrained at each step to impose symmetry.
\subsection{Computing the irreps from input representations}
\label{sec:iterative}
For some groups, the computation of the generators $X$ can become a very involved task. However in most applications, the data itself is already given in a form of a representation. One approach proposed by \citep{Finzi2021:mlp} is to not work in the space of irreps but the space of polynomials of the input representation. This approach has the advantage of requiring little previous knowledge of the group. However it is also much less efficient than using irreps. One alternative way is to consider polynomials of the input representation, that are reducible and then compute the block diagonalisation to project down to irreps subspace. One can then work directly as polynomials in this subspace and compute Clebsch--Gordan coefficients numerically. We provide routines in \texttt{lie-nn} to carry out these operations from any given input representation.

\subsection{Details of numerical experiments}

\subsubsection{Jet Tagging}
\label{sec:jet_tagging_appendix}
\paragraph{Dataset}
The dataset~\citep{Kasieczka:2019dbj} was generated using a \texttt{Pythia}, \texttt{Delphes}, and \texttt{FastJet} (using cuts for the jet's kinematics on $\Delta\eta=2$, $R=0.8$) to simulate the response of the ATLAS detector at the Large Hadron Collider (LHC).
The dataset is released under the "Creative Commons Attribution 4.0" license.
The entire dataset contains 2 millions jets with a 60/20/20 for training, validation, and testing balanced splits.

\paragraph{Model}
The model uses \textbf{3 layers} of the $G$-MACE architecture to generate the Lorentz group equivariant representation of each jet. For the 1 particle basis, we use a product of radial features on the Minkowski distances, and $SO(1,3)$ spherical harmonics.
The radial features are computing by passing a logarithmic radial basis as in \citep{Bogatskiy:2022czk} into a $[64,64,64, 512]$ MLP using SiLU nonlinearities on the outputs of the hidden layers.
The internal representations used are $(0,0)$ and $(1,1)$. We use 72 channels for each representation. 
For the embedding, and readout out, we use similar achitectures to LorentzNet.

\paragraph{Training}
Models were trained on an NVIDIA A100 GPU in single GPU training. Typical training time for the dataset is up to 72 hours.
Models were trained with AMSGrad variant of Adam, with default parameters of $\beta_{1} = 0.9$, $\beta_{2} =
0.999$, and $\epsilon = 10^{-8}$. We used a learning rate of $0.0035$ and a batch size of 64. 
The model was trained for 80 epochs with 2 epochs of linear learning rate warmup and followed by a phase of cosine annealing LR scheduling.

\subsubsection{3D shape recognition}
\label{sec:point_appendix}
\paragraph{Dataset}
ModelNet10~\citep{WU2015:modelnet} is a synthetic 3D object point clouds dataset containing 4,899 pre-aligned shapes from 10 categories. The dataset is split into 3,991 $(80\%)$ shapes for training and 908 $(20\%)$ shapes for testing. We were unable to find a license. 

\paragraph{Model}
The model uses a three-layer encoder architecture following the PointNet++ one. We use an encoder of the full point cloud into sub-point clouds of sizes $[1024, 256, 128]$. Each PointNet layer maps a point cloud of size $N^{t}$ to one of size $N^{t+1}$. We compute the node features as the sum of the PointNet output and the MACE output,
\begin{equation}
    h^{(t+1)} = \text{PointNet}(xyz^{(t)}, h^{(t)}) + \text{MACE}(xyz^{(t)}, h^{(t)})
\end{equation}

\paragraph{Training}
Models were trained on an NVIDIA A100 GPU in single GPU training. The typical training time for the dataset is up to 12 hours.
Models were trained with the AMSGrad variant of Adam, with default parameters of $\beta_{1} = 0.9$, $\beta_{2} =
0.999$, and $\epsilon = 10^{-8}$.

\subsection{Limitations and Future Work}

The spectrum of potential applications of the present method is very large. In this paper, we focus on a subset of applications that have known benchmarks and baselines. A broader range of groups is implemented in the lie-nn library. Future work should focus on applying this architecture to tasks with domain-specific knowledge.

\end{document}